\documentclass[10pt,twocolumn,letterpaper]{article}

\usepackage[pagenumbers]{cvpr} 

\usepackage{graphicx}
\usepackage{amsmath}
\usepackage{amssymb}
\usepackage{booktabs}

\usepackage{xcolor}
\usepackage{colortbl}
\usepackage{times}
\usepackage{epsfig}
\usepackage{soul}
\usepackage{float}
\usepackage{adjustbox}

\definecolor{oblue}{HTML}{0A3161}
\definecolor{ored}{HTML}{B31942}

\colorlet{citecolor}{ored}
\colorlet{urlcolor}{ored}

\usepackage[pagebackref,breaklinks,colorlinks,allcolors=ored]{hyperref}

\usepackage[capitalize]{cleveref}
\crefname{section}{Sec.}{Secs.}
\Crefname{section}{Section}{Sections}
\Crefname{table}{Table}{Tables}
\crefname{table}{Tab.}{Tabs.}
\Crefname{appendix}{Appendix}{Appendices}

\DeclareMathAlphabet\mathbfcal{OMS}{cmsy}{b}{n}

\newcommand{\natten}{$\mathcal{N}\hspace{-0.2em}ATTEN$}

\newcommand{\nattensim}{\natten{}Sim}
\newcommand{\nattensimbold}{$\mathbfcal{N}\hspace{-0.2em}\mathbf{A}$\textbf{Sim}}

\newcommand{\duo}[2]{#1 \texttimes{} #2}
\newcommand{\trio}[3]{#1 \texttimes{} #2 \texttimes{} #3}

\newcommand{\Speedup}[1]{#1\texttimes{}}

\newcommand{\bigO}{\mathcal{O}}

\urldef{\solurl}\url{https://docs.nvidia.com/nsight-compute/ProfilingGuide/index.html#roofline-charts}
\urldef{\tmaurl}\url{https://docs.nvidia.com/cuda/hopper-tuning-guide/index.html#tensor-memory-accelerator}

\def\cvprPaperID{*****} 
\def\confName{CVPR}
\def\confYear{2025}

\begin{document}

\title{Generalized Neighborhood Attention:\\
Multi-dimensional Sparse Attention at the Speed of Light}

\author{
    Ali Hassani\textsuperscript{1,2},
    Fengzhe Zhou\textsuperscript{1},
    Aditya Kane\textsuperscript{1},
    Jiannan Huang\textsuperscript{1},\\
    Chieh-Yun Chen\textsuperscript{1},
    Min Shi\textsuperscript{1},
    Steven Walton\textsuperscript{1},
    Markus Hoehnerbach\textsuperscript{2},\\
    Vijay Thakkar\textsuperscript{1,2},
    Michael Isaev\textsuperscript{2},
    Qinsheng Zhang\textsuperscript{2},
    Bing Xu\textsuperscript{2},
    Haicheng Wu\textsuperscript{2},\\
    Wen-mei Hwu\textsuperscript{2,3},
    Ming-Yu Liu\textsuperscript{2},
    Humphrey Shi\textsuperscript{1,3} \\
{\small
\textsuperscript{1}Georgia Tech,
\textsuperscript{2}NVIDIA,
\textsuperscript{3}UIUC
}\\
{\small \textbf{\url{https://github.com/SHI-Labs/NATTEN}}}
}

\twocolumn[{
    \vspace{-7mm}
    \maketitle
    \vspace{-5.5mm}
    \centering
    \includegraphics[width=\textwidth]{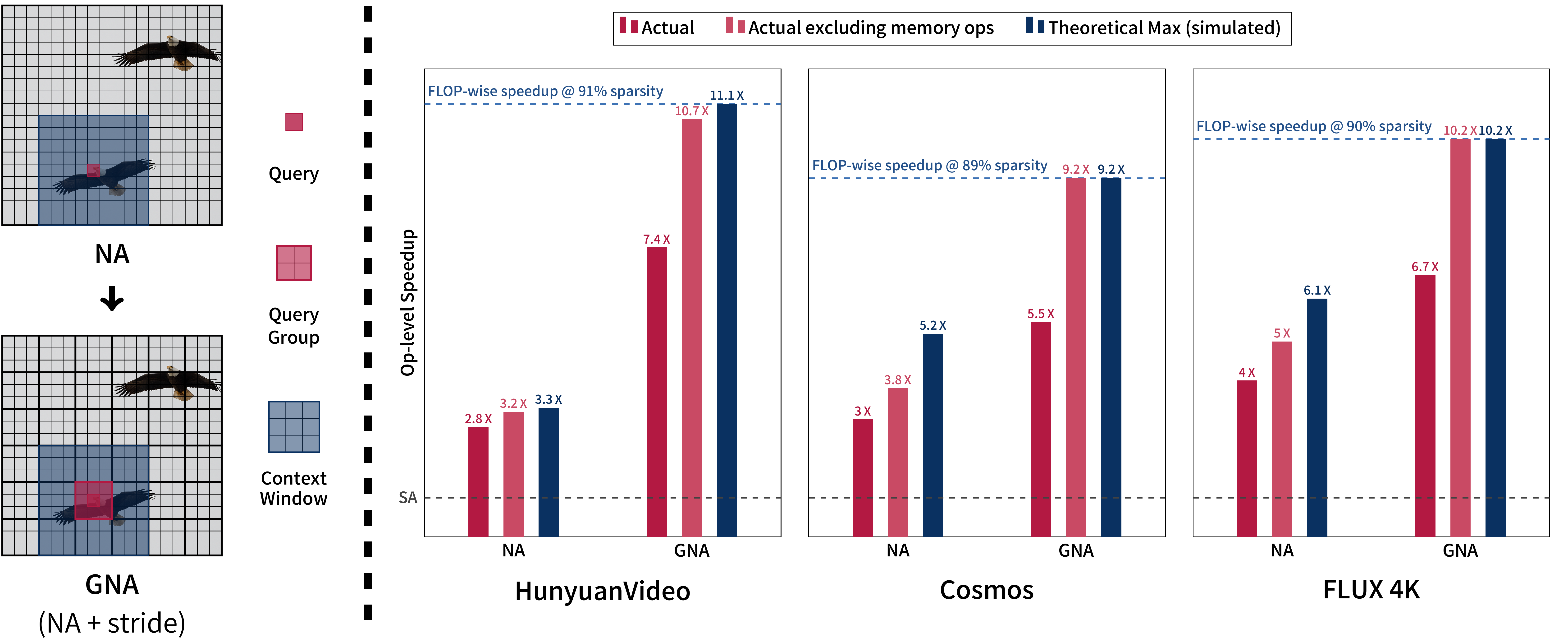}
    \vspace{-4.5mm}
    \captionsetup{type=figure}
    \captionof{figure}{
        \textbf{Generalized Neighborhood Attention} adds a new ``stride'' parameter to neighborhood attention, which
        introduces a delay in the sliding window by grouping queries together
        and forcing them to share their context window. This increases density in matrix
        multiplications, while maintaining overall sparsity, leading to speedups more proportional to savings in FLOPs.
        Our Blackwell kernel can as a result realize the \textbf{maximum speedup theoretically possible} in some cases,
        with respect to both FLOPs and simulation.
    }
    \label{fig:teaser}
    \vspace{5mm}
}]

\begin{abstract}
Many sparse attention mechanisms such as Neighborhood Attention have typically failed to consistently
deliver speedup over the self attention baseline.
This is largely due to the level of complexity in attention infrastructure, and the rapid evolution
of AI hardware architecture.
At the same time, many state-of-the-art foundational models, particularly in computer vision, are heavily bound by attention,
and need reliable sparsity to escape the $\bigO (n^2)$ complexity.
In this paper, we study a class of promising sparse attention mechanisms that
focus on locality, and aim to develop a better analytical model of their performance improvements.
We first introduce Generalized Neighborhood Attention (GNA), which can describe sliding window,
strided sliding window, and blocked attention.
We then consider possible design choices in implementing these approaches, and create a simulator
that can provide much more realistic speedup upper bounds for any given setting.
Finally, we implement GNA on top of a state-of-the-art fused multi-headed attention (FMHA) kernel designed for the
NVIDIA Blackwell architecture in CUTLASS.
Our implementation can fully realize the \textbf{maximum speedup theoretically possible} in many perfectly block-sparse cases,
and achieves an effective utilization of 1.3 petaFLOPs/second in FP16.
In addition, we plug various GNA configurations into off-the-shelf generative models, such as Cosmos-7B,
HunyuanVideo, and FLUX, and show that it can deliver 28\% to 46\% end-to-end speedup on B200 without any fine-tuning.
We will open source our simulator and Blackwell kernels directly through the \natten{} project.
\end{abstract}

\section{Introduction}
Fast sparse attention has been long sought-after
\cite{parmar2018image,ramachandran2019stand,child2019generating,beltagy2020longformer,zaheer2020big,vaswani2021scaling,qiu2020blockwise},
but rarely without complications.
Infrastructure continues to be a challenge for attention in general, as implementations of attention rarely
come close to adequately utilizing the computational power of modern GPUs, at least compared to dense
linear algebra primitives, such as generalized matrix-matrix multiplications (GEMMs), which 
typically utilize around 80\% of peak FLOPs/second.
The most successful example to date is Flash Attention 3~\cite{shah2024flashattention}, utilizing up to
75\% of the peak FLOPs/second in the NVIDIA Hopper architecture with half precision.
However, lower precision still trails behind, and
with every new architecture, and changes in the programming model, new challenges in infrastructure arise.
Sparse approaches that require changes to the core attention implementation have therefore lagged behind.
One example is approaches that specifically target sparsity in attention weights,
such as sliding window~\cite{parmar2018image,ramachandran2019stand,beltagy2020longformer,hassani2023neighborhood}
and blocked attention~\cite{vaswani2021scaling,liu2021swin},
which can only accelerate computation by utilizing block-sparsity.
Block-sparsity is primarily comprised of skipping tiles of computation that are fully masked in some
pre-defined attention mask.
Implementing block-sparsity can be non-trivial, and at times comes with significant overhead
that can undo performance gains, assuming the base implementation is already the state of the art.
As a result, these approaches usually leave some performance on the table, with lower utilization than dense attention.

Despite these challenges, sliding window and blocked attention have been successfully built into
large language models~\cite{jiang2023mistral,cohere2025command,meta2025llama4},
where block-sparsity and fine-grained masking are concerned with a single-dimensional token layout.
This makes their implementation significantly simpler compared to
applications such as images and videos, which step into 2-D and 3-D token layouts.
As a direct result of this, implementations focused on multi-dimensional token layouts typically fall behind
even further.
One such example is Fused Neighborhood Attention (FNA)~\cite{hassani2024faster}, which implements multi-dimensional
sliding window attention, but suffers a considerable performance gap due to overheads that worsen with
higher dimensionality, namely software predication and fine-grained masking.
At the same time, FNA targets the NVIDIA Ampere architecture, which results in even larger gap compared to
the state-of-the-art on newer architectures, such as Hopper.
While some successful implementations of multi-dimensional local attention have recently appeared~\cite{zhang2025fast},
they are not as flexible in terms of pattern and parameters, offering only fully block-sparse masks,
and are tied to very specific use cases.

Presently a variety of frameworks that provide linear algebra and tensor layout primitives exist, such as
CuTe and CULTASS~\cite{thakkar2023cutlass}, Triton~\cite{tillet2019triton},
and compilers specifically designed for block-sparse attention, such as Flex Attention~\cite{dong2024flex}.
Despite the amount of impact this has had on the advancement of research, wildly different implementations
of different approaches make it increasingly difficult to analyze the exact performance differences in these
multi-dimensional sparse attention methods, and identify their root causes.

\begin{figure*}[ht!]
    \centering
    \includegraphics[width=1.0\linewidth]{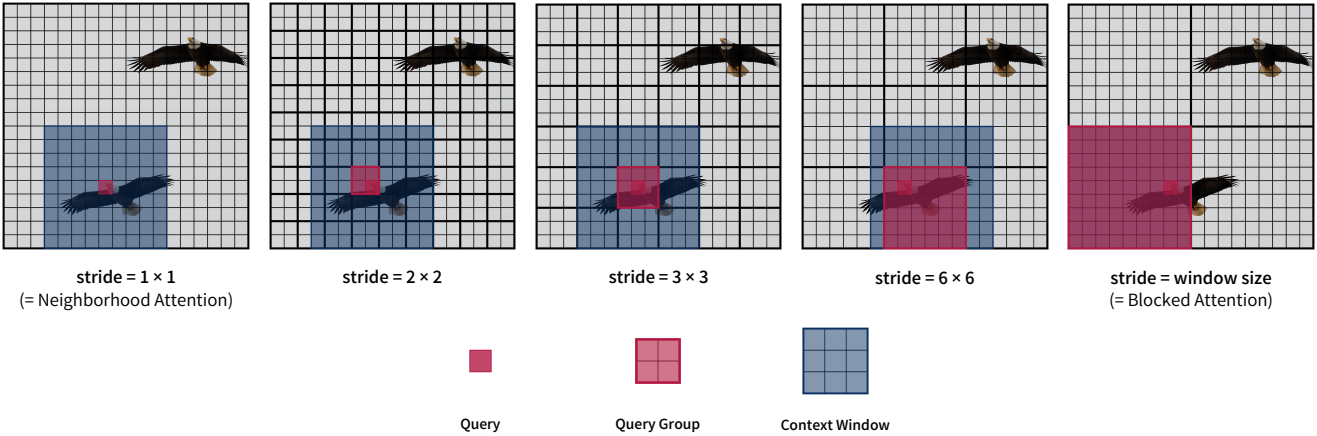}
    \caption{
        Generalized Neighborhood Attention allows customizable ``delay steps'' in the sliding window pattern through
        the new \textbf{stride} parameter. Stride can take any positive integer value smaller than or equal to
        window size. Stride of 1 is equivalent to standard neighborhood attention. When stride is equal to window
        size, it is equivalent to blocked attention (a.k.a. Window Self Attention).
    }
    \label{fig:gna-strides}
\end{figure*}

To that end, we propose Generalized Neighborhood Attention (GNA), an extension to Neighborhood Attention
(NA)~\cite{hassani2023neighborhood}, itself a sliding window approach, which aims to unify many existing
approaches under one definition.
GNA can offer a tradeoff space between efficiency, and quality and inductive biases (i.e. translational equivariance).
Approaches implementable with GNA can be classified into sliding window~\cite{parmar2018image,ramachandran2019stand,hassani2023neighborhood},
strided sliding window~\cite{vaswani2021scaling,zhang2025fast}, and blocked~\cite{vaswani2021scaling,liu2021swin} attention.
We revisit the problem of sliding window vs strided sliding window and blocked attention discussed in HaloNet~\cite{vaswani2021scaling},
but with focus on speedup as opposed to memory operations.
An illustration of different GNA patterns is presented in \cref{fig:gna-strides}.

We further create a simulation tool, which can compute the upper-bound speedup achievable by any approach defined
under GNA, under different use cases, and implementation design choices. Through abstracting away implementation details,
and defining speedup with tile/block granularity as opposed to FLOP/FMA granularity, we can finally fairly compare these different
approaches analytically.

Finally, we implement GNA on top of CUTLASS's attention kernel for the NVIDIA Blackwell architecture,
which is one of the best-performing choices available today, and show that we can successfully realize up
to 100\% of the FLOP-wise efficiency in many GNA configurations.

Although attention itself typically does not come close to standard GEMMs in
terms of Speed-of-Light (SoL) performance~\footnote{\solurl},
given the negligible performance overhead of our implementation,
we believe our methodology can be a recipe for sparse and local attention methods to finally catch up to the
performance of standard FMHA implementations.

We identify variants of GNA that can exhibit perfect speedup, and introduce both them, and standard Neighborhood Attention
into generative and foundation models, such as Cosmos~\cite{agarwal2025cosmos}, HunyuanVideo~\cite{kong2024hunyuanvideo}, and
Flux~\cite{flux2024}.
We show that using our Blackwell kernel, which we have directly integrated into the \natten{} project, we can accelerate these
models without fine-tuning up to 63\% end-to-end.
Beyond improving and accelerating \natten{}, and supporting more sparse attention patterns, we hope to provide a useful set of
tools, such as our simulator, to help researchers find the most useful configurations for their models, and easily ablate over
a useful set of parameters, instead of being lost at the outrageously large parameter space that GNA has to offer.
All of our work will be open-sourced.

\section{Related Works}
Attention has been widely called out for being a computationally intensive operation, as a result of its quadratic
complexity with respect to number of tokens. This has often been one of the motivations behind research into sparse forms of attention.
Dot-product attention, the most widely used form of attention, consists primarily of two matrix multiplications, which
means it can enjoy many forms of sparsity which exist for matrix multiplies, such as structured sparsity~\cite{chen2023dynamic},
and low-rank approximations~\cite{wang2020linformer,chen2021scatterbrain}.
Sparsity can also be introduced into attention by choosing coarser targets of sparsity, which is sometimes application-specific.
For example, some approaches are designed specifically for introducing sparsity into LLM inference workloads, where KV caching
is a necessity, and therefore calls for some form of sparsity or compression~\cite{shazeer2019fast,ainslie2023gqa,liu2024deepseek}.
Approaches such as Token Merging~\cite{bolya2023token} attempt to directly reduce the number of both query and context tokens
together, and have been shown to be effective in certain vision tasks.

Sparse Attention approaches can also be classified into static and dynamic approaches.
Dynamic approaches~\cite{roy2021efficient,bolya2023token,chen2023dynamic} can be more effective without further fine-tuning,
whereas static approaches~\cite{parmar2018image,chen2021scatterbrain,ramachandran2019stand,child2019generating,beltagy2020longformer}
are more likely to achieve better speedup~\cite{jiang2023mistral,cohere2025command,liu2021swin}.
Some approaches can be classified as hybrids~\cite{xu2025xattention,bolya2023token,xi2025sparse}, where statistics
gathered at runtime guide re-layout of computation, and then use static methods such as block-sparsity~\cite{qiu2020blockwise}.
Some may even use entirely dense computations~\cite{liu2021swin,bolya2023token}.
Some, such as Native Sparse Attention~\cite{yuan2025native}, combine multiple sparse approaches, both static and dynamic.

In this paper, we specifically focus on static methods, where the target of sparsity is the attention weight matrix itself,
and therefore the token coordinate space determines whether or not a region / weight is masked.
Local attention is the most prominent example, but variants of local attention where global context is introduced through
non-contiguous context windows~\cite{beltagy2020longformer,hassani2022dilated} also fall into this category.
The success of local approaches~\cite{liu2021swin,kirillov2023segment,li2022exploring,jain2023oneformer,crowson2024scalable,du2025weathermesh}
can be attributed to the fact that they require little to no change to the Transformer~\cite{vaswani2017attention} architecture,
as well as spatial locality being a bias that exists in nature.
Past~\cite{jiang2023mistral} and recent~\cite{cohere2025command,meta2025llama4} large language models
also adopt local methods into their architecture in addition to standard self attention.

\subsection{(Locally) Sparse Attention}
Over the years, many proposed introducing locality into attention, some for the inductive biases, some
as a means to achieve subquadratic complexity, and some for both.
In this section, we summarize these approaches into three categories, which are described below.

\paragraph{Sliding Window Attention.}
Some of the earliest works proposing this were Image Transformer~\cite{parmar2018image} and
Stand-Alone Self-Attention (SASA)~\cite{ramachandran2019stand},
where they specifically designed a form of attention sharing some of the inductive biases present in convolution.
The sliding window pattern itself is quite similar to, and sometimes inspired by, convolution.
However, they did not become widely adopted at first, particularly in vision, due to a lack of infrastructure.
Ramachandran et al.~\cite{ramachandran2019stand} stated that while their model achieved superior accuracy
to a comparable CNN, it required more resources and time to be trained.
The same concept, but in a single-dimensional causal domain, was later used in language, in works such as
Longformer~\cite{beltagy2020longformer}, and extended to a ``dilated'' form, which can introduce global context
with the same complexity as standard local attention.
Years later, neighborhood attention (NA)~\cite{hassani2023neighborhood} revisited this concept for hierarchical vision
transformers. The key difference between SASA and NA is in the handling of corner cases in the coordinate space.
SASA employed a zero-padded feature map, similar to padding in convolution.
NA ``reflects'' the window back, so that every query is guaranteed to attend to a fixed
number of context tokens, regardless of its position in the coordinate space and window size.
This behavior also guarantees that NA numerically matches self attention when window size is equal to input size.
Similar to Longformer, but again in the context of vision, a dilated form of NA was later introduced~\cite{hassani2022dilated}.
The combination of standard and dilated NA in a hierarchical vision transformer surpassed the original model in accuracy
across various tasks~\cite{hassani2022dilated,walton2022stylenat}, without incurring any additional cost.

\paragraph{Strided Sliding Window Attention.}
Also referred to as blocked local attention~\cite{vaswani2021scaling}, this approach effectively introduces
a delay step into sliding window attention.
This was originally motivated by the fact that typical implementations of sliding window attention~\cite{parmar2018image,ramachandran2019stand}
required explicit memory operations that extract sliding windows from context, which can quickly undo any savings in
computation, in addition to growing the memory footprint.
This form of sliding window attention has an extreme case, called Blocked Attention, where stride is as large
as the window size, which results in non-overlapping local windows.
More recently, this approach was revisited in Sliding Tile Attention (STA)~\cite{zhang2025fast}, where
the objective was achieving perfectly block-sparse masks as a means to minimize masked, and therefore wasted FLOPs 
in sliding window methods~\cite{parmar2018image,ramachandran2019stand,hassani2023neighborhood}.
We will further clarify this approach in \cref{subsec:curse-of-multi-dim}.

\paragraph{Blocked Attention.}
This approach effectively partitions the query and context set, and performs self attention on each partition
independently (and in parallel) to the rest. In addition to HaloNet~\cite{vaswani2021scaling},
Window Self Attention (WSA) from Swin Transformer~\cite{liu2021swin} is another instance of blocked attention.
Blocked attention is easy to implement, and embarrassingly parallel along blocks, but this comes at the cost
of no cross-block interactions. This can be remedied by introducing global attention, or varying window (block) sizes
across layers (Swin's Shifted WSA), or introducing convolution.
This approach has been adopted in many vision models such as Segment Anything~\cite{kirillov2023segment}, and
ViTDet~\cite{li2022exploring}.

\subsection{(Sparse) Attention Infrastructure}
Sliding window attention, specifically in the context of vision, was commonly considered
inefficient~\cite{vaswani2021scaling,liu2021swin}, but that was predicated on the assumption that any
form of transformation to context tokens (keys and values) must be explicit in global memory.
One of the key contributions of Neighborhood Attention Transformer~\cite{hassani2023neighborhood} was
a set of naive CUDA kernels that simply computed the vector-matrix multiplication problem without
such explicit copies in global memory. These kernels were packaged as a PyTorch extension, called \natten{}.
However, around the same time as the initial release, implementation of attention in general was about to undergo
a massive change.

Until 2022~\cite{rabe2021self,dao2022flashattention}, most implementations of dot-product attention, especially those
not tied to specific inference use cases, were implemented with two matrix multiplications, with
the softmax operator in between. In most deep learning frameworks, the former typically targets
General Matrix-Matrix Multiplication (GEMM) routines in powerful dense linear algebra packages,
i.e. cuBLAS and CUTLASS~\cite{thakkar2023cutlass} for NVIDIA GPUs, and these routines typically offer great performance,
usually up to 80\% of the peak FLOPs/second utilization.
The issue with this approach however is that the size of the intermediary matrix, the attention weight matrix, grows
quadratically, resulting in a quadratic memory footprint, and by extension quadratic number of memory operations in
both GEMMs. As a result, this implementation becomes heavily limited by memory bandwidth, which on most modern GPUs
is orders of magnitude smaller than computation power (FLOPs/second).
Flash Attention~\cite{dao2022flashattention} showed that by fusing the two GEMMs and softmax into
the same kernel, and utilizing online softmax~\cite{milakov2018online,jang2019mnnfast}, we can continuously accumulate context in SRAM
without ever fully realizing the $\bigO(n^2)$ weight matrix.
This approach is also referred to as Fused Multi-headed Attention (FMHA).
Flash Attention 2~\cite{dao2023flashattention} later improved upon the original, and is the state-of-the-art for the
NVIDIA Ampere architecture.
Flash Attention 3~\cite{shah2024flashattention} extended the approach to the NVIDIA Hopper architecture by using the new
programming model and hardware features through CuTe and CUTLASS~\cite{thakkar2023cutlass}.

Due to the fact that self attention is the baseline for all sparse attention, and that baseline is improved significantly
with FMHA, sparse attention approaches have had to follow suit.
Mistral-7B~\cite{jiang2023mistral} was one of the first models to use 1-D sliding window attention by directly implementing it
as a block-sparse mask in state-of-the-art implementations, such as Flash Attention 2~\cite{dao2023flashattention}.
More recently, language models such as Command A~\cite{cohere2025command} have also adopted this approach, and
use sliding window attention together with global self attention.

However, implementations such as \natten{} faced additional challenges, due to the additional burden of dealing with
multi-dimensional layouts of tokens, which makes efficient block-sparsity for them non-trivial.

\begin{figure}[ht!]
    \centering
    \includegraphics[width=1.0\linewidth]{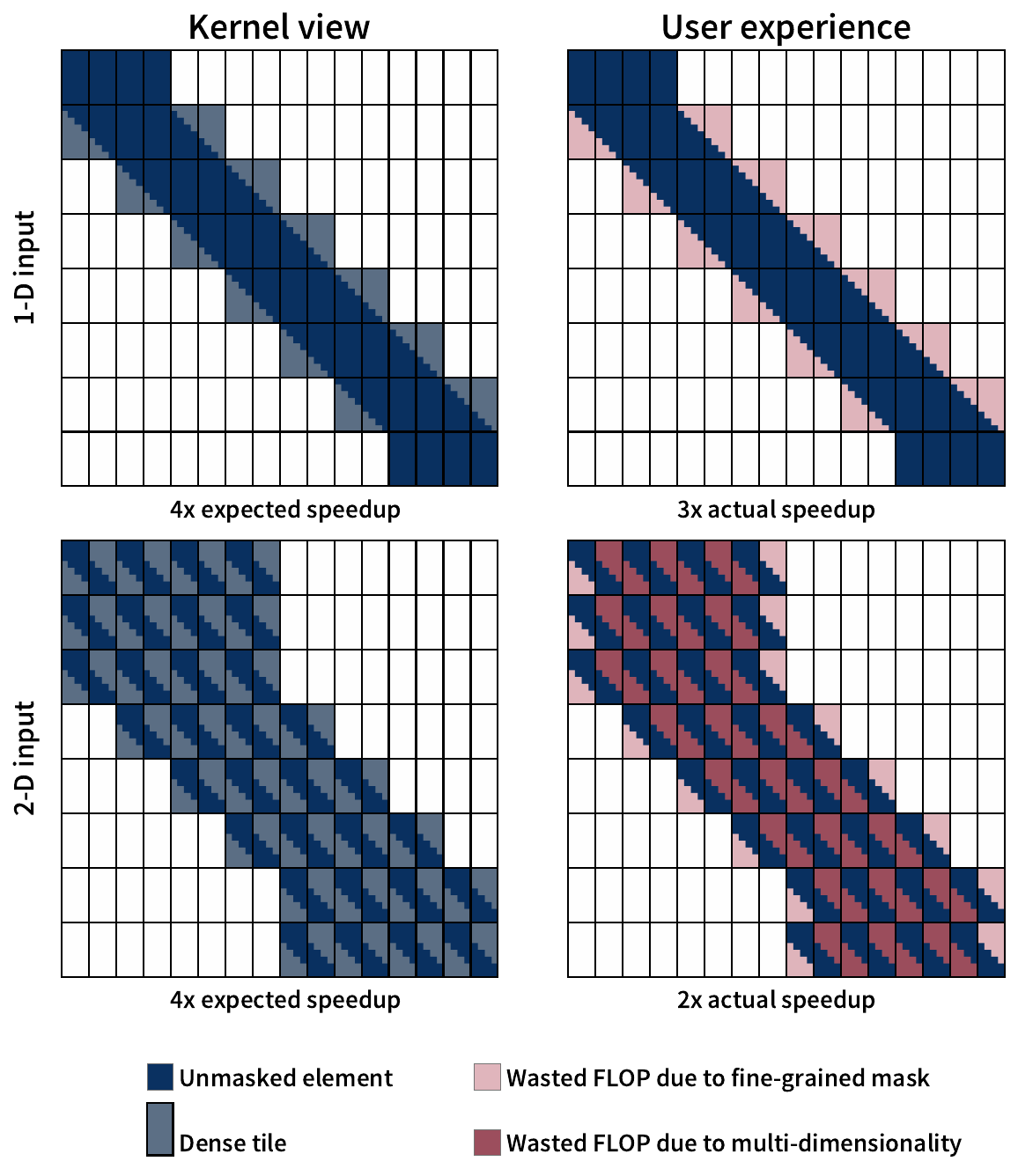}
    \caption{
        Curse of \textbf{multi-dimensionality}: single-dimensional tiling opens up sparsity in multi-dimensional
        layouts of tokens to more wasted computation (FLOPs that are still computed but masked prior to softmax).
        While many fine-grained attention masks, even 1-D sliding window attention (top), can still have some FLOPs masked
        due to the fact that the vector-matrix multiplies are packed into matrix-matrix multiplies,
        masked FLOPs due to multi-dimensionality can be much more significant (bottom).
        Note that the single-dimensional case is bi-directional and not causal for better comparison to the multi-dimensional
        case.
    }
    \label{fig:curse-of-multi-dim}
\end{figure}

\subsection{Curse of Multi-Dimensionality}
\label{subsec:curse-of-multi-dim}
Studies involving sparsity in attention weights have one key, and often undiscussed, difference between LLM-focused applications
and vision applications, which is the multi-dimensional layout of visual data (images, videos).
We consider this an additional burden, as design choices available for efficient implementations are
limited, and often with noticeable overhead.
To illustrate this issue, we refer readers to \cref{fig:curse-of-multi-dim}.
When considering for instance a 2-dimensional layout of tokens (i.e. a feature map in an image model), if the FMHA
implementation employs single-dimensional tiling over the typical layout of these tokens in memory, the potential
for ``wasted compute'' increases, and according to the tile size.

Fused Neighborhood Attention (FNA)~\cite{hassani2024faster} proposed solving this by employing multi-dimensional tiling
which converts the GEMMs in the FMHA into an instance of tensor contractions.
This implementation naturally improved naive kernels in \natten{} by a significant margin, but it can also be
greatly limited by the overhead of software predication as a direct result of multi-dimensional tiling.
While hardware predication, through components like NVIDIA's Tensor Memory Accelerator (TMA)~\footnote{\tmaurl},
can likely minimize this overhead, they can also impact certain design choices critical to achieving optimal speedup,
such as dynamic KV tiling.

A more easily implementable alternative to multi-dimensional tiling is simulating the behavior through re-layout of tokens.
One of the earliest demonstrations of this was in the implementation of FNA in Flex Attention~\cite{dong2024flex}\footnote{Dubbed ``Tiled NATTEN''.}.
Flex Attention is a PyTorch API that evaluates user-defined attention masks, and compiles them into block-sparse Triton kernels.
This approach adds a non-avoidable overhead from the additional memory operations required for the re-layout.
It also still requires some modification of the original attention kernel, but far fewer changes compared to fused multi-dimensional
tiling.
In the case of Flex Attention, this can be handled by translating 1-D coordinates into coordinates in the new layout of tokens
directly in the user-defined mask.
Sliding Tile Attention (STA)~\cite{zhang2025fast} attempts to further minimize FLOPs wasted due to fine-grained masking
(see \cref{fig:curse-of-multi-dim}), and proposes defining neighborhood / sliding window attention on tiles / blocks instead
of individual tokens, with the tile / block size matching that of the underlying FMHA kernel.
This closely resembles strided sliding window approaches such as HaloNet~\cite{vaswani2021scaling},
but is instead motivated by computational density, and therefore speedup, instead of the cost of memory operations,
which are non-existent with block-sparse FMHA kernels.
STA's implementation employs re-layout of tokens from Flex Attention, instead of on-the-fly mutli-dimensional tiling,
and successfully outperforms FNA given the same FLOP-wise sparsity.
It is noteworthy that the implementation is specific to the Hopper architecture, whereas FNA targets Ampere tensor cores.
Aside from this, the key limitation of STA is that like Blocked Attention, it assumes query and context tensors are always
tiled according to the same tile size.
This assumption is not guaranteed to hold in practice, and even if some implementations support it, there is no guarantee that
such configurations achieve optimal performance for the given use case.

Given that there are many approaches to local sparse attention, and each with various design choices in their
implementation, and the various properties and performance levels of different hardware architectures,
we aim to disambiguate their differences and similarities.
To do so, we first unify them under a new framework, which we call
``\textbf{Generalized Neighborhood Attention}'' (GNA).
Under this framework, we create unified implementations for sliding window~\cite{parmar2018image,ramachandran2019stand} / neighborhood~\cite{hassani2023neighborhood} attention,
strided sliding window attention~\cite{vaswani2021scaling},
blocked attention~\cite{vaswani2021scaling,liu2021swin}, and
approaches that fall in between, including special cases such as STA~\cite{zhang2025fast}.
This includes extending \natten{} to support all of these methods.

We further build an analytical tool, called \nattensim{}, which computes a more fine-grained upper-bound speedup
for these approaches under different implementation design choices.
This is helpful for both optimizing speed or accuracy given an existing implementation,
as well as for investigating whether or not a new design choice / implementation is justified based on its implications on
end-to-end speedup.

Finally, we create a new FNA kernel for the Blackwell architecture inspired by the original FNA~\cite{hassani2024faster},
and based on CUTLASS's Blackwell FMHA kernel~\cite{thakkar2023cutlass}, and show that in many cases, we can completely
match the analytical speedup, both with respect to FLOPs and with respect to \nattensim{}.

\section{Methodology}

\begin{figure*}[ht!]
    \centering
    \includegraphics[width=1.0\linewidth]{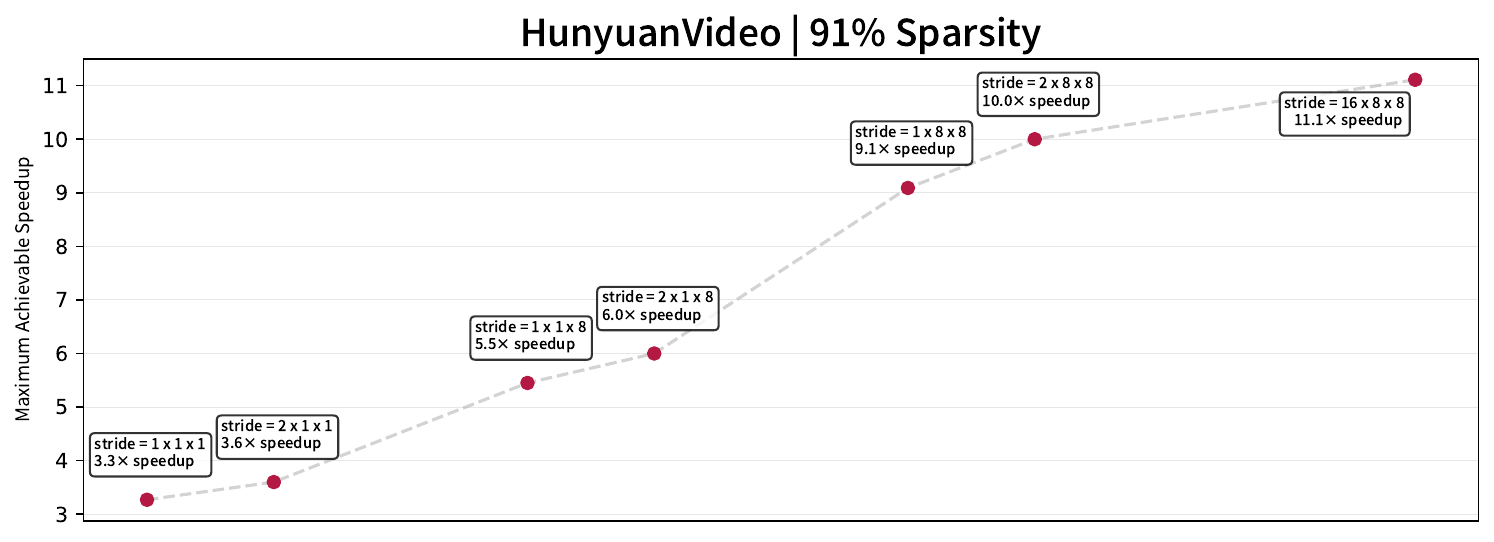}
    \caption{
        Sweep of different stride values and their analytical speedup according to \nattensim{},
        with a window size of \trio{18}{24}{24} ($\approx 91\%$ sparsity).
        The simulation assumes a Q tile shape of \trio{4}{8}{8}, and KV tile shape of
        \trio{2}{8}{8}, a combination supported by our Blackwell kernel.
        Standard neighborhood attention (stride 1) is limited by a \Speedup{3.3} speedup, while some larger strides
        can improve upon that, and eventually cross \Speedup{9} speedup with a stride of \duo{8}{8} across the
        spatial axes.
        With some larger strides along the temporal axis, it can reach perfect block-sparsity, and yield
        a speedup of \Speedup{11.1}, which is equivalent to its FLOP-wise speedup.
    }
    \label{fig:tilesim-hunyuan}
\end{figure*}

In this section, we define Generalized Neighborhood Attention, which introduces a new dimension to the neighborhood attention
family of static attention patterns, and is solely aimed at providing a tradeoff space between accuracy and translational equivariance,
and efficiency.
We then describe design choices in implementations of such methods, which is one of our motivations for creating our analytical tool,
\nattensim{}.
We describe how the simulator functions, and how it can be used to guide the process of selecting parameters for GNA based on information
about the use case, design choices, and even hardware-specific details.
We finally move on to our implementation for the Blackwell architecture, which was in part motivated by studies conducted with our analytical
tool. This implementation is based on one of the best-performing FMHA kernels available for the Blackwell architecture, and was
done within the powerful CuTe and CUTLASS~\cite{thakkar2023cutlass} framework.

\subsection{Generalized Neighborhood Attention}
Neighborhood Attention (NA)~\cite{hassani2023neighborhood}, and more generally, sliding window attention patterns are fine-grained
masks that allow each query token to have a unique context window of fixed size, which is determined based on its coordinate.
The effect of this is similar to that of convolution, where a filter of a fixed size is applied to sliding window regions, and
contracted to a single point in the output. The key difference between the approaches is that the filter in convolution is static
and typically learned through gradient descent, whereas in sliding window attention, the ``filter'' is dynamic, and based on context.
Previous works have extended NA to support dilated windows~\cite{hassani2022dilated}, mimicking dilation in convolution, as well
as causal masking~\cite{hassani2024faster}, which can be useful for video applications, where an atomic spatio-temporal
attention mask may be required.

In this work, we start with relaxing the definition of standard NA to allow for even values for window size.
NA was originally defined on odd values for window size, as the goal is for the query to be perfectly centered in its neighborhood.
We do so by splitting \textbf{window size} into two: window size left, and window size right.
In the standard definition of NA, both are equivalent to the floor of window size divided by two.
In the case of even-sized windows, we can choose either side to be larger by 1.
By default, we choose this to be the left side, meaning given window size 8, a non-corner token attends to
4 tokens on its left side, itself, and 3 tokens on its right side.

We further add a fourth parameter to NA, which we call ``\textbf{stride}'', as it can be reminiscent of stride in convolution.
As in convolution, stride adds a delay step to the sliding windows in NA, meaning a stride of 1 would be equivalent to standard NA,
moving the sliding window for every query token, while a stride of 2 moves the sliding window for every two query tokens.
Another way of viewing this is that stride \textbf{groups queries together}, and forces them to share the same neighborhood (context window).
Queries in each group statically elect a leader query and attend to its neighborhood, as defined by standard NA.
The choice of the leader, similar to window left and window right, can be user-defined, but the default choice in GNA
is the center-most query in the group, for consistency with the original NA work.
If the group is evenly-sized, it would be biased towards the right side
\footnote{The reason for this choice is that query group leader being biased by one to the right can cancel out
window size left being biased by one on the left in some cases, which can enable perfect block-sparsity.}.
Stride can take any positive integer value less than or equal to window size.
Strides larger than window size introduce ``holes'' in attention, where some context tokens are never attended to,
and are therefore disallowed.
We present a visualization of different strides for a fixed window size in \cref{fig:gna-strides}.
Strided NA exhibits similar behavior as blocked local attention in HaloNet~\cite{vaswani2021scaling}, with
their ``block size'' parameter introducing the same delay step effect into sliding window
attention~\cite{parmar2018image,ramachandran2019stand} as stride does in neighborhood attention~\cite{hassani2023neighborhood}.
Another interesting property in both is that when stride is equal to window size, the pattern will be equivalent
to (fully) blocked attention, also known as Window Self Attention (WSA)~\cite{liu2021swin}.
In some cases, Strided NA can also implement STA~\cite{zhang2025fast}, but we note this is only guaranteed
when certain assumptions are made with respect to the underlying implementation. We will clarify this further
in \cref{sec:nattensim,sec:blackwell-fna}.

Stride being a new key parameter in the NA family of sparse attention patterns is largely motivated by the fact that
\textit{sliding window attention cannot achieve speedup that is perfectly proportional to its sparsity ratio}.
This is simply because sliding window attention is a vector-matrix multiplication
\footnote{Generalized Matrix-Vector Multiply (GEMV) in BLAS.} problem commonly implemented with matrix-matrix multiplies (GEMMs)
~\cite{hassani2024faster} and masking over dot products.
Masking ensures correctness but at the same time it is wasting work (FLOPs) already done.
On the other hand, stride forces queries in the same group to share their entire neighborhood,
which aligns the boundaries of their windows together. This means stride can bridge the gap between
sliding window approaches~\cite{parmar2018image,ramachandran2019stand,hassani2023neighborhood}, and
blocked approaches~\cite{vaswani2021scaling,liu2021swin,zhang2025fast}.
The advantage of blocked approaches is that given certain assumptions, they can mainly perform dense computation,
and do not require fine-grained masking of attention weights, and by extension do not ``waste'' FLOPs.
More specifically, any approach that is guaranteed to be perfectly block-sparse~\cite{qiu2020blockwise}
by definition only involves dense computation, and can lead to speedups more proportional to savings in FLOPs.
As a result, we aim to answer questions such as: how much work is exactly wasted, and do larger strides guarantee
fewer wasted FLOPs.
In order to do so, we created an analytical tool for simulating the behavior of common implementations for such methods,
and use it to compute how many \textit{tiles} of work is each combination saving.

\subsection{Analytical tool for GNA}
\label{sec:nattensim}
Stride extends an already vast parameter space. This, along with the different design choices and configurations in implementation,
specifically that of FMHA kernels, makes it difficult to find useful parameters for end-users that offer the best tradeoff.
We therefore created an analytical tool called \nattensim{}, which can shed light on exactly that.
\nattensim{} computes the number of context (KV) tiles visited by an implementation, taking into account various design choices
such as dynamic vs static KV tiling, 1-D vs multi-dimensional tiling, and of course tile shapes for query (Q) and context (KV) tensors.

\paragraph{Design choice: tile sizes.}
Among the most important configurations for any GEMM or GEMM-based kernel, tile sizes directly determine how the workload
is divided up and scheduled among workers.
In most FMHA kernels, which fuse together two GEMMs, they are forced to share most of their tile sizes,
but with a permutation.
In the first GEMM, query tensor (Q) is tiled along the query sequence mode with some tile size $T_Q$,
key tensor (K) is tiled along the context sequence mode with some tile size $T_K$.
The shared head dim mode, which is the contracting mode, is tiled by some tile size $T_D$.
In the second GEMM, the lefthandside operand is a tile of attention weights of shape $T_Q \times T_K$,
which means the righthandside operand, tile from value tensor (V), must match the tile size along the contracting dimension,
which is now the context sequence mode. Therefore, K and V take the same tile size, which we will henceforth refer to as
$T_{KV}$. The head dim mode of V is also typically tiled by $T_D$.

Tile sizes $T_Q$ and $T_{KV}$ therefore directly affect the number of FLOPs masked (wasted).
Many factors can determine valid choices for these tile sizes, such as the amount of SRAM available,
number of stages in the GEMM pipeline, layout transformations required for the operands,
and of course shapes of the underlying matrix multiply-and-accumulates (MMAs), or Tensor Core instruction shapes
in the case of NVIDIA GPUs.
With modern architectures such as Hopper and Blackwell, and in the case of GEMMs, there are many choices
for tile sizes available, which exhibit different performance levels on different problem shapes.
One key limitation in works such as STA~\cite{zhang2025fast} is that the methodology assumes $T_Q$ is always
equivalent to $T_{KV}$, thus limiting the number of choices.
This is while performant FMHA kernels are not guaranteed to always provide the same level of flexibility when it
comes to picking tile sizes as in standard dense GEMMs.

\paragraph{Design choice: Single-dimensional vs multi-dimensional tiling.}
Most FMHA kernels assume a single-dimensional layout of tokens, which given attention's permutation equivariance
is logical. This however creates a challenge for cases where tokens assume a multi-dimensional layout, such as
visual models where tokens represent patches of pixels in images and videos.
This opens up those applications to more wasted FLOPs, as illustrated in \cref{fig:curse-of-multi-dim}.
A natural fix is to tile in multiple dimensions and with respect to the original multi-dimensional layout,
which essentially converts the GEMM problem into a tensor contraction (GETT\footnote{Generalized Tensor-Tensor Contraction (GETT).}).
This solution was employed by FNA~\cite{hassani2024faster}, where $T_Q$ and $T_{KV}$ in the base FMHA kernel
were re-interpreted as multi-dimensional tiles (i.e. $T_Q = 64 \rightarrow T_Q = 8 \times 8$).
One downside to this is that this can introduce a significant overhead due to additional software predication logic,
and given that the kernel was designed for the Ampere architecture, hardware predication was not an option.
A practical solution to this is taking multi-dimensional tiling out of the kernel and instead implementing it
as a re-layout operation.
This solution was employed by Flex Attention~\cite{dong2024flex} in their implementation of multi-dimensional
NA masks (referred to as ``Tiled NATTEN'').
The only potential downside to this approach is the unavoidable fixed cost of memory operations, which is independent
of the level of sparsity, and only a function of the size of the Q, K, V, and output tensors.
We dub this approach \textbf{token permutation}, as it is mainly comprised of a re-layout of the token space,
and is agnostic to batch, heads, and head dim.

\begin{figure*}[ht!]
    \centering
    \newcommand{\imgwidth}{0.495}
    \begin{subfigure}{\imgwidth\textwidth}
        \includegraphics[width=1.0\linewidth]{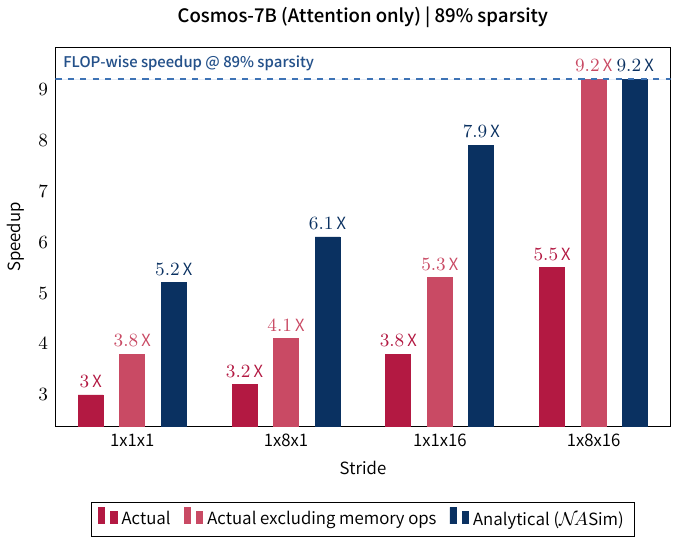}
        \caption{
            Cosmos-7B use case with 89\% sparse GNA (window size \trio{16}{24}{16}).
        }
    \end{subfigure}
    \begin{subfigure}{\imgwidth\textwidth}
        \includegraphics[width=1.0\linewidth]{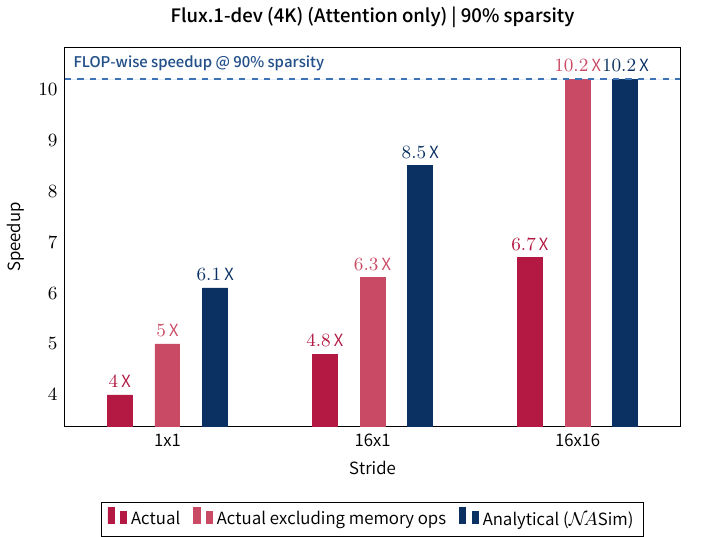}
        \caption{
            FLUX.1-dev (4K) use case with 90\% sparse GNA (window size \duo{80}{80}).
        }
    \end{subfigure}
    \\
    \vspace{2.5mm}
    \renewcommand{\imgwidth}{0.72}
    \begin{subfigure}{\imgwidth\textwidth}
        \includegraphics[width=1.0\linewidth]{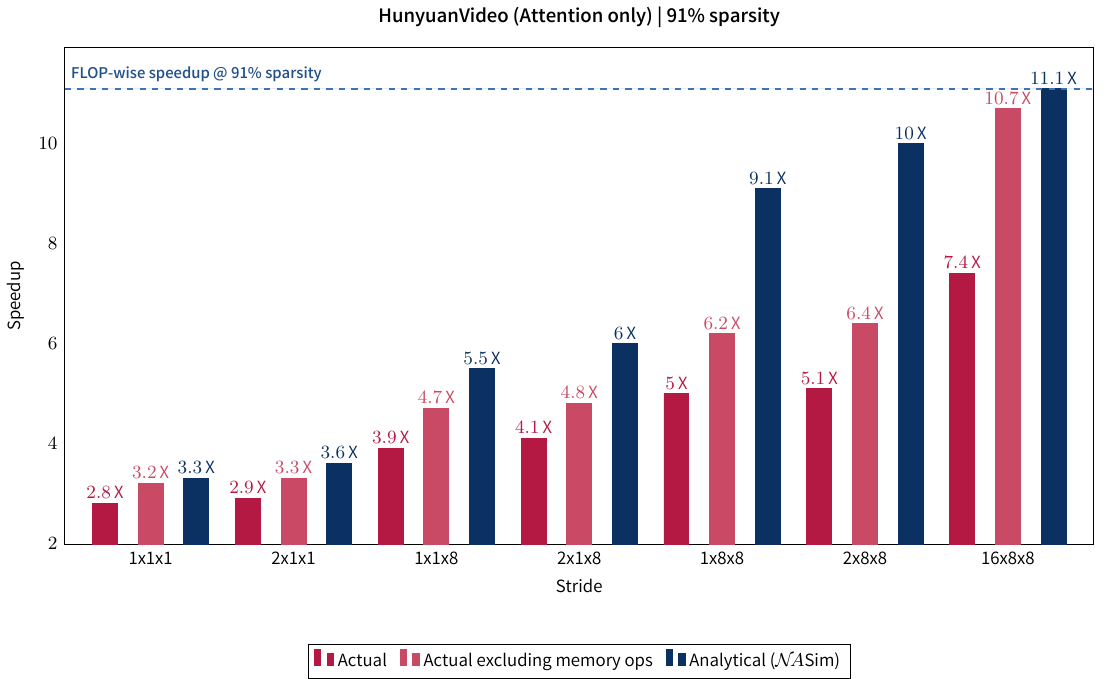}
        \caption{
            HunyuanVideo use case with 91\% sparse GNA (window size \trio{18}{24}{24}).
        }
    \end{subfigure}
    \caption{
        Operation-level (attention only) speedups on Cosmos, FLUX, and HunyuanVideo with $\approx 90\%$
        sparsity through GNA.
        Analytical speedup is according to \nattensim{}, and actual speedups are measured by
        running on B200.
        Note that with the perfectly block-sparse strides, our
        kernel can come very close to or fully match the full the analytical speedup,
        but is limited by the naive implementation of the the memory operation (token permute).
    }
    \label{fig:blackwell-fna-speedups}
\end{figure*}

\paragraph{Design choice: KV tiling.}
Tiling is typically static, and in many block-sparse FMHA kernels, static KV tiles are either visited, or skipped, according
to the mask definition.
However, some implementations, such as FNA~\cite{hassani2024faster}, first slice out the region in the KV
token space that would be required to be visited, and dynamically tile the region.
This can save some additional computation, and result in minimal wasted FLOPs possible.
On the other hand, this approach is not always realizable, especially in designs that rely on hardware predication.
For instance, the Hopper Tensor Memory Accelerator (TMA) requires determining the parameters of tiled copies
(tensor maps) prior to kernel launch. While on-the-fly modification/replacement of tensor maps is possible, it is
not without overhead.

\paragraph{Use cases.}
Problem shape (layout of tokens), window size, dilation, stride, and causal masking are all user-specified,
and play a role in determining computational savings in terms of tiles.

\paragraph{Tiling simulation.}
\nattensim{}'s primary goal is to simulate how a given use case is tiled according to design choices in the
implementation. Through basic operations on coordinates, and by using an exact definition of the core GNA
mask, \nattensim{} computes the coordinates of each KV tile visited by each Q tile.
If we consider the worst case of all Q tiles (maximum number of KV tiles), we can compute a more realistic
and fine-grained upper bound speedup than FLOP-wise speedup, with respect to self attention for each use case.
For instance, a perfectly blocks-sparse mask and sliding window attention mask with 90\% sparsity both have
a FLOP-wise upper-bound speedup of \Speedup{10} ($\frac{1}{1 - 90\%}$), but the latter never get away with
performing exactly $\frac{1}{10}$ of the FLOPs unless implemented as a vector-matrix multiplication.

With \nattensim{}, we can achieve the following:
\begin{enumerate}
    \item Compare different design choices, and pick the one best trading off implementation difficulty and speedup.
    \item Find perfectly block-sparse cases, where no fine-grained masking is required, and simulated speedup matches FLOP-wise speedup,
        as illustrated in \cref{fig:tilesim-hunyuan}.
        Under dynamic KV tiling, any setting in which $T_{KV}$ evenly divides window size,
        and $T_{Q}$ evenly divides stride achieves this.
        Under static KV tiling, it is less straightforward, and some simulation may be required to find those points.
    \item Predict end-to-end speedup upper bounds for any given model under a specific set of parameters and design choices.
\end{enumerate}

\subsection{Implementation for the Blackwell architecture}
\label{sec:blackwell-fna}
We start off with the CUTLASS FMHA kernel for Blackwell~\cite{thakkar2023cutlass}, which can achieve up to 1.2 petaFLOPs/s
with FP16 and up to 1.7 petaFLOPs/s with FP8 precision (w/ per-tensor scaling).
While our implementation specifically focuses on the Blackwell architecture, the design choices in the implementation are
primarily architecture-agnostic.
This implementation closely follows the original FNA kernel~\cite{hassani2024faster}, with the exception of
taking multi-dimensional tiling outside the kernel, and instead relying on token permutation.
This also forces static KV tiling instead of dynamic KV tiling in the original FNA.
We chose this design for the following reasons:

\begin{enumerate}
    \item If using the TMA for data movement, static KV tiling is the only choice.
    
    \item Fusing multi-dimensional tiling into existing FMHA kernels can break too many assumptions made with respect to
    the sequence mode, even with CuTe facilitating layout transformations and interfacing with the TMA.

    \item We are not leaving much performance on the table, as long as we are not
    limited by the memory transfer time from token permutation and reverse permutation.
    For example, considering use cases from Cosmos-7B~\cite{agarwal2025cosmos} and HunyuanVideo~\cite{kong2024hunyuanvideo},
    this would be only 6.9 and 10.5GB respectively. If we utilize even half of the 8TB/s HBM bandwidth of a single B200,
    this would be 1.9\% and 1\% of the FMHA time, which would only limit very large levels of sparsity.
    
    \item In many cases, token permutation and reverse permutation can be done only once: permute before the first
    transformer layer, and reverse after the last layer. Most transformer architectures are equivariant to permutation,
    and this holds true for both ViT~\cite{dosovitskiy2020image} and DiT~\cite{peebles2023scalable}, which are prominent in vision.
    
    \item Fusing additional context tokens is trivial with token permutation.
    Certain models, such as Hunyuan~\cite{kong2024hunyuanvideo} and Flux~\cite{flux2024} cross attend visual tokens with
    text tokens. \natten{} has usually supported those scenarios by launching an additional FMHA kernel, and ``merging''
    the two outputs using their logsumexp.
    However, we implement this feature within the same kernel, allowing some KV tokens to take a completely different layout
    and mask.
\end{enumerate}

Since we implement multi-dimensional tiling, spatial locality allows us to define visited KV tiles as the range between
the last KV tile coordinate required by the last query in the Q tile, and the first KV tile coordinate required
by the first query in the Q tile. Most of the kernel remains agnostic to the multi-dimensional tiling, and
only the producer and softmax warps take this into account. Producer warp simply maps iteration index to
the multi-dimensional tile coordinate, and then to the static 1-D tile index, which directly interfaces with
the indexing of TMA loads through CuTe.
Softmax warp(s) likewise have to map the 1-D Q and KV coordinates back into the original multi-dimensional
coordinates, and apply the fine-grained GNA mask. However, given the overhead of masking, we also implement
predicates for perfectly block-sparse cases, and for any additional KV tokens, which can in some cases completely
eliminate our implementation overhead.

In the current iteration, we implement token permutation as a copy operation through PyTorch directly.
Problem shapes that are not evenly divisible are manually padded, output tensor is cropped after
reverse token permutation, and kernel handles predication for KV padding.
In future versions, we hope to also improve upon token permutation, as the current solution typically
utilizes approximately 1/8th of the memory bandwidth.
The kernel and the additional memory operations, and potential padding, is directly integrated into \natten{},
and exposed via the typical \verb|na{1,2,3}d| API.

\section{Experiments}
In order to evaluate GNA's applicability and performance,
we carefully selected applications with multi-dimensional layouts of tokens that are suitable for
sparse attention.
Our primary criterion for choosing applications is whether or not at least 40\% of their end-to-end workload
is self attention specifically.
Our final candidates are Cosmos~\cite{agarwal2025cosmos} (World Foundation Model),
HunyuanVideo~\cite{kong2024hunyuanvideo} (video generation), and FLUX~\cite{flux2024} (image generation).
We additionally note that FLUX only meets the criterion when generating resolutions higher than 2K.
We present the workload distribution for those models in \cref{tab:workload-pctages}.

\begin{table}[t]
    \centering
    \resizebox{0.475\textwidth}{!}{
    \begin{tabular}{l|ccc}
        \toprule
        \textbf{Use case}        & \textbf{\% SA in} & \textbf{\% DiT in}    & \textbf{\% SA in}     \\
                                 & \textbf{DiT}      & \textbf{E2E workload} & \textbf{E2E workload} \\
        \midrule
        \textbf{Cosmos-7B}       &  58.7\%             & $>$ 99\%                  & 58.7\%         \\
        \textbf{HunyuanVideo}    &  65.4\%             & 92.8\%                    & 60.7\%         \\
        \textbf{FLUX.1-dev (4K)} &  56.8\%             & 91.2\%                    & 51.8\%         \\
        \bottomrule
    \end{tabular}
    }
    \caption{
        Workload distribution with respect to self attention in Cosmos-7B, HunyuanVideo, and FLUX.1-dev.
        Measurements were done on a single B200 and without any additional performance optimizations, and
        using the original FP16/BF16 precision.
    }
    \label{tab:workload-pctages}
\end{table}

\begin{table}[t]
    \centering
    \resizebox{0.475\textwidth}{!}{
    \begin{tabular}{lll|rr|r}
        \toprule
        \textbf{Window} & \textbf{Stride} & \textbf{\# SA} & \multicolumn{3}{c}{\textbf{E2E Speedup} $\uparrow$} \\
        \textbf{Size}   &                 & \textbf{Steps} & \multicolumn{2}{c|}{\textbf{Analytical}}  & \textbf{Actual}              \\
                        &                 &                & \textbf{FLOP-wise}  & \nattensimbold{} &              \\
        \midrule
        \rowcolor[HTML]{EFEFEF} \multicolumn{6}{l}{\textbf{56\% sparsity}} \\
        \trio{16}{32}{48} & \trio{1}{1}{1}   & 0        & \Speedup{1.50}     & \Speedup{1.35}     & \Speedup{1.18} \\
        \trio{16}{32}{48} & \trio{1}{8}{1}   & 0        & \Speedup{1.50}     & \Speedup{1.42}     & \Speedup{1.20} \\
        \trio{16}{32}{48} & \trio{1}{1}{16}  & 0        & \Speedup{1.50}     & \Speedup{1.42}     & \Speedup{1.21} \\
        \trio{16}{32}{48} & \trio{1}{8}{16}  & 0        & \Speedup{1.50}     & \Speedup{1.50}     & \Speedup{1.46} \\
        \midrule
        \trio{16}{32}{48} & \trio{1}{1}{1}   & 12       & \Speedup{1.28}     & \Speedup{1.21}     & \Speedup{1.11} \\
        \trio{16}{32}{48} & \trio{1}{8}{1}   & 12       & \Speedup{1.28}     & \Speedup{1.24}     & \Speedup{1.12} \\
        \trio{16}{32}{48} & \trio{1}{1}{16}  & 12       & \Speedup{1.28}     & \Speedup{1.24}     & \Speedup{1.13} \\
        \trio{16}{32}{48} & \trio{1}{8}{16}  & 12       & \Speedup{1.28}     & \Speedup{1.28}     & \Speedup{1.26} \\
        \midrule
        \rowcolor[HTML]{EFEFEF} \multicolumn{6}{l}{\textbf{89\% sparsity}} \\
        \trio{16}{24}{16} & \trio{1}{1}{1}   & 0        & \Speedup{2.10}     & \Speedup{1.90}     & \Speedup{1.76} \\
        \trio{16}{24}{16} & \trio{1}{8}{1}   & 0        & \Speedup{2.10}     & \Speedup{1.96}     & \Speedup{1.79} \\
        \trio{16}{24}{16} & \trio{1}{1}{16}  & 0        & \Speedup{2.10}     & \Speedup{2.05}     & \Speedup{1.88} \\
        \trio{16}{24}{16} & \trio{1}{8}{16}  & 0        & \Speedup{2.10}     & \Speedup{2.10}     & \Speedup{2.05} \\
        \midrule
        \trio{16}{24}{16} & \trio{1}{1}{1}   & 12       & \Speedup{1.52}     & \Speedup{1.45}     & \Speedup{1.40} \\
        \trio{16}{24}{16} & \trio{1}{8}{1}   & 12       & \Speedup{1.52}     & \Speedup{1.48}     & \Speedup{1.40} \\
        \trio{16}{24}{16} & \trio{1}{1}{16}  & 12       & \Speedup{1.52}     & \Speedup{1.51}     & \Speedup{1.44} \\
        \trio{16}{24}{16} & \trio{1}{8}{16}  & 12       & \Speedup{1.52}     & \Speedup{1.52}     & \Speedup{1.50} \\
        \bottomrule
    \end{tabular}
    }
    \caption{
    Cosmos-7B end-to-end speedups from GNA under different sparsity levels and strides.
    We report both analytical speedups based on \nattensim{}, as well as actual
    speedups on B200.
    We also report two settings: one with all diffusion steps done with GNA, and the other
    with the first 12 steps done with self attention. The former can be more representative
    of cases where the model can be fine-tuned with GNA, while the latter is more applicable for
    off-the-shelf applications.
    }
    \label{tab:cosmos-e2e-speedups}
\end{table}

\begin{table}[t]
    \centering
    \resizebox{0.475\textwidth}{!}{
    \begin{tabular}{lll|rr|r}
        \toprule
        \textbf{Window} & \textbf{Stride} & \textbf{\# SA} & \multicolumn{3}{c}{\textbf{E2E Speedup} $\uparrow$} \\
        \textbf{Size}   &                 & \textbf{Steps} & \multicolumn{2}{c|}{\textbf{Analytical}}  & \textbf{Actual}              \\
                        &                 &                & \textbf{FLOP-wise}  & \nattensimbold{} &              \\
        \midrule
        \rowcolor[HTML]{EFEFEF} \multicolumn{6}{l}{\textbf{58\% sparsity}} \\
        \trio{30}{40}{40} & \trio{1}{1}{1}   & 0        & \Speedup{1.55} & \Speedup{1.21}     & \Speedup{1.15} \\
        \trio{30}{40}{40} & \trio{1}{1}{8}   & 0        & \Speedup{1.55} & \Speedup{1.44}     & \Speedup{1.26} \\
        \trio{30}{40}{40} & \trio{1}{32}{8}  & 0        & \Speedup{1.55} & \Speedup{1.55}     & \Speedup{1.53} \\
        \midrule
        \trio{30}{40}{40} & \trio{1}{1}{1}   & 15       & \Speedup{1.33} & \Speedup{1.14}     & \Speedup{1.08} \\
        \trio{30}{40}{40} & \trio{1}{1}{8}   & 15       & \Speedup{1.33} & \Speedup{1.27}     & \Speedup{1.15} \\
        \trio{30}{40}{40} & \trio{1}{32}{8}  & 15       & \Speedup{1.33} & \Speedup{1.33}     & \Speedup{1.30} \\
        \midrule
        \rowcolor[HTML]{EFEFEF} \multicolumn{6}{l}{\textbf{91\% sparsity}} \\
        \trio{18}{24}{24} & \trio{1}{1}{1}   & 0         & \Speedup{2.23} & \Speedup{1.73}     & \textbf{\Speedup{1.73}} \\
        \trio{18}{24}{24} & \trio{2}{1}{1}   & 0         & \Speedup{2.23} & \Speedup{1.78}     & \Speedup{1.77} \\
        \trio{18}{24}{24} & \trio{1}{1}{8}   & 0         & \Speedup{2.23} & \Speedup{1.99}     & \Speedup{1.95} \\
        \trio{18}{24}{24} & \trio{2}{1}{8}   & 0         & \Speedup{2.23} & \Speedup{2.02}     & \Speedup{1.98} \\
        \trio{18}{24}{24} & \trio{1}{8}{8}   & 0         & \Speedup{2.23} & \Speedup{2.17}     & \Speedup{2.09} \\
        \trio{18}{24}{24} & \trio{2}{8}{8}   & 0         & \Speedup{2.23} & \Speedup{2.20}     & \Speedup{2.11} \\
        \trio{18}{24}{24} & \trio{16}{8}{8}  & 0         & \Speedup{2.23} & \Speedup{2.23}     & \textbf{\Speedup{2.23}} \\
        \midrule
        \trio{18}{24}{24} & \trio{1}{1}{1}   & 15        & \Speedup{1.63} & \Speedup{1.42}     & \textbf{\Speedup{1.42}} \\
        \trio{18}{24}{24} & \trio{2}{1}{1}   & 15        & \Speedup{1.63} & \Speedup{1.44}     & \textbf{\Speedup{1.44}} \\
        \trio{18}{24}{24} & \trio{1}{1}{8}   & 15        & \Speedup{1.63} & \Speedup{1.53}     & \Speedup{1.51} \\
        \trio{18}{24}{24} & \trio{2}{1}{8}   & 15        & \Speedup{1.63} & \Speedup{1.55}     & \Speedup{1.54} \\
        \trio{18}{24}{24} & \trio{1}{8}{8}   & 15        & \Speedup{1.63} & \Speedup{1.61}     & \Speedup{1.58} \\
        \trio{18}{24}{24} & \trio{2}{8}{8}   & 15        & \Speedup{1.63} & \Speedup{1.62}     & \Speedup{1.59} \\
        \trio{18}{24}{24} & \trio{16}{8}{8}  & 15        & \Speedup{1.63} & \Speedup{1.63}     & \textbf{\Speedup{1.63}} \\
        \bottomrule
    \end{tabular}
    }
    \caption{
    HunyuanVideo end-to-end speedups from GNA under different sparsity levels and strides.
    We report both analytical speedups based on \nattensim{}, as well as actual
    speedups on B200.
    We also report two settings: one with all diffusion steps done with GNA, and the other
    with the first 15 steps done with self attention. The former can be more representative
    of cases where the model can be fine-tuned with GNA, while the latter is more applicable for
    off-the-shelf applications.
    }
    \label{tab:hunyuan-e2e-speedups}
\end{table}

\subsection{GNA Performance}
We first consider cases under which these applications introduce roughly 90\% of sparsity into attention,
and run their problem shapes through \nattensim{}, sweeping over all possible stride values.
For each use case, we selected Q and KV tile shapes ($T_Q$ and $T_{KV}$) according to the shape of the token layout
(feature map size).
We chose window sizes that are evenly divisible by the KV tile shape ($T_{KV}$), as this can result
in perfectly block-sparse forms of GNA, which waste no FLOPs and can potentially achieve perfect speedup.
A similar observation was made in STA~\cite{zhang2025fast}, with the exception that STA does not differentiate
between $T_Q$ and $T_{KV}$, and that their step size parameter is fixed.

In addition, we prune the results of the simulator according to a simple but useful filter.
We use the product of stride values as a measure for grouping different strides, and eliminate any
configuration in which larger strides do not improve performance compared to smaller strides.
For instance, if any case with stride $>$ 1 achieves worse analytical speedup over NA itself (stride 1),
we do not report it.
This is especially helpful, since larger strides trade off translational equivariance and potentially quality
for potentially better performance, and if the latter is not realized under some setting, that configuration
is unlikely to have any advantage over others.

We present one instance of results obtained through \nattensim{} in \cref{fig:tilesim-hunyuan}, which
is on the HunyuanVideo use case with approximately 90\% sparsity. Unlike most others,
in this particular use case we have many choices with varying upper-bound speedups, any of which
may offer the best accuracy-efficiency trade-off if further fine-tuned or trained with.
We also report the actual performance of our Blackwell kernel on the three models with 90\%
sparsity, and compare against the analytical upper bound reported by \nattensim{} in \cref{fig:blackwell-fna-speedups}.
Two observations can immediately be made:
1. In perfectly block-sparse cases, our kernel either fully or almost fully realizes the analytical speedup, which is also
the FLOP-wise speedup (maximum achievable).
This comes at no surprise, since the use of the fine-grained masking in the softmax stage is conditioned on whether or not
it is required.
2. In the case of Hunyuan, even the standard Neighborhood Attention comes very close to realizing its full analytical speedup.
However, the cost of fine-grained masking can still bear some non-negligible overhead in cases that are not perfectly block-sparse.

\noindent We report end-to-end measures of speedup in \cref{tab:flux-e2e-speedups,tab:cosmos-e2e-speedups,tab:hunyuan-e2e-speedups}.
In addition, we present qualitative results, as well as some quantitative benchmarks on these use cases in the next two subsections.

\begin{table}[t]
    \centering
     \resizebox{0.475\textwidth}{!}{
    \begin{tabular}{lll|rr|r}
        \toprule
        \textbf{Window} & \textbf{Stride} & \textbf{\# SA} & \multicolumn{3}{c}{\textbf{E2E Speedup} $\uparrow$} \\
        \textbf{Size}   &                 & \textbf{Steps} & \multicolumn{2}{c|}{\textbf{Analytical}}  & \textbf{Actual}              \\
                        &                 &                & \textbf{FLOP-wise}  & \nattensimbold{} &              \\
        \midrule
        \duo{80}{80} & \duo{1}{1}   & 0        & \Speedup{1.88}     & \Speedup{1.76}     & \Speedup{1.65} \\
        \duo{80}{80} & \duo{16}{1}  & 0        & \Speedup{1.88}     & \Speedup{1.84}     & \Speedup{1.72} \\
        \duo{80}{80} & \duo{16}{16} & 0        & \Speedup{1.88}     & \Speedup{1.88}     & \Speedup{1.82} \\
        \midrule                                
        \duo{80}{80} & \duo{1}{1}   & 9        & \Speedup{1.46}     & \Speedup{1.42}     & \Speedup{1.37} \\
        \duo{80}{80} & \duo{16}{1}  & 9        & \Speedup{1.46}     & \Speedup{1.45}     & \Speedup{1.40} \\
        \duo{80}{80} & \duo{16}{16} & 9        & \Speedup{1.46}     & \Speedup{1.46}     & \Speedup{1.45} \\
        \bottomrule
    \end{tabular}
    }
    \caption{
    FLUX-1.dev with URAE~\cite{yu2025ultra} end-to-end speedups from GNA under different strides.
    We report both analytical speedups based on \nattensim{}, as well as actual
    speedups on B200.
    We also report two settings: one with all diffusion steps done with GNA, and the other
    with the first 9 steps done with self attention. The former can be more representative
    of cases where the model can be fine-tuned with GNA, while the latter is more applicable for
    off-the-shelf applications.
    }
    \label{tab:flux-e2e-speedups}
\end{table}

\subsection{Video Generation}

\begin{table*}[ht!]
    \centering
    \resizebox{1.0\textwidth}{!}{
    \begin{tabular}{llll||rrr||c|rr|r}
        \toprule
        \textbf{Method}         & \textbf{Window Size} & \textbf{Stride} & \textbf{Attention} & \multicolumn{3}{c||}{\textbf{VBench}} & \textbf{Runtime $\downarrow$} & \multicolumn{3}{c}{\textbf{End-to-End Speedup $\uparrow$}}\\
                                &                      &                 & \textbf{Sparsity} & \textbf{Total} $\uparrow$ & \textbf{Quality} $\uparrow$ & \textbf{Semantic} $\uparrow$ &
                \textbf{on B200} & \multicolumn{2}{c|}{\textbf{Analytical}} & \textbf{Actual}\\
        &&&&&&& \textbf{(seconds)} & \textbf{FLOP-wise} &\nattensimbold{} & \\
        \midrule
        \textbf{SA} (baseline)  & -                 & -                 & 0.0\%    & 83.08\% & 85.01\% & 75.35\% & 628 & & &  \\
        \midrule
        \rowcolor[HTML]{EFEFEF} \multicolumn{11}{l}{\textbf{\textit{As reported in STA~\cite{zhang2025fast}}}} \\
        \textbf{Tiled NATTEN} (= NA)& \trio{30}{41}{41} & -             & 58.3\%   & 82.69\% & 84.61\% & 75.00\% & - & \Speedup{1.55} & - & -            \\
        \textbf{Swin} (= WSA)   & \trio{30}{40}{40} & -                 & 58.3\%   & 77.53\% & 78.84\% & 72.28\% & - & \Speedup{1.55} & - & -            \\
        \textbf{STA}            & \trio{30}{40}{40} & -                 & 58.3\%   & 82.46\% & 84.63\% & 73.83\% & - & \Speedup{1.55} & - & -            \\
        
        \midrule
        \textbf{STA}            & \trio{18}{24}{24} & -                 & 91.0\%   & 80.58\% & 81.47\% & 77.03\% & - & \Speedup{2.23} & - & -            \\
        \textbf{STA w/ training}& \trio{18}{24}{24} & -                 & 91.0\%   & 82.62\% & 84.76\% & 74.05\% & - & \Speedup{2.23} & - & -            \\
        \midrule
        \rowcolor[HTML]{EFEFEF} \multicolumn{11}{l}{\textbf{\textit{Ours}}} \\
        \textbf{GNA} (= NA)     & \trio{30}{40}{40} & \trio{1}{1}{1}    & 58.3\%   & 83.24\% & 84.70\% & 77.42\% & 546 & \Speedup{1.55} & \Speedup{1.21} & \Speedup{1.15} \\
        \textbf{GNA} (= WSA)    & \trio{30}{40}{40} & \trio{30}{40}{40} & 58.3\%   & 82.25\% & 83.11\% & 78.83\% & 491 & \Speedup{1.55} & \Speedup{1.44} & \Speedup{1.28}    \\
        \textbf{GNA}            & \trio{30}{40}{40} & \trio{1}{32}{8}   & 58.3\%   & 83.40\% & 84.77\% & 77.92\% & 411 & \Speedup{1.55} & \Speedup{1.55} & \Speedup{1.53} \\
        \midrule
        \textbf{GNA} (= NA)     & \trio{18}{24}{24} & \trio{1}{1}{1}    & 91.0\%   & 82.36\% & 83.03\% & 79.68\% & 359 & \Speedup{2.23} & \Speedup{1.73} & \textbf{\Speedup{1.73}} \\
        \textbf{GNA} (= WSA)    & \trio{18}{24}{24} & \trio{18}{24}{24} & 91.0\%   & 80.25\% & 80.97\% & 77.38\% & 314 & \Speedup{2.23} & \Speedup{2.06} & \Speedup{2.00} \\
        \textbf{GNA}            & \trio{18}{24}{24} & \trio{16}{8}{8}   & 91.0\%   & 82.04\% & 82.62\% & 79.69\% & 277 & \Speedup{2.23} & \Speedup{2.23} & \textbf{\Speedup{2.23}} \\
        \bottomrule
    \end{tabular}
    }
    \caption{
        HunyuanVideo VBench performance across different GNA configurations.
        Following STA~\cite{zhang2025fast}, these configurations do not have any self attention steps.
        We do not report model FLOPs, as it is a poor measure of efficiency, and instead report two analytical speedups: one
        based on FLOPs, which would be the same across methods for any given attention sparsity, and another based on
        \nattensim{}.
        We also report actual speedup based on GNA runtimes, which use our Blackwell FNA kernel.
        However, note that the above runtimes can vary $\pm$ 3.5\%.
        We do not report end-to-end speedups from STA, as our study was done on the Blackwell architecture,
        and STA's implementation was specific to the Hopper architecture, and the difference in workload distribution
        would skew the comparison. We further note it is not possible to run kernels that target Hopper Tensor Cores on any
        architecture other than Hopper.
        In addition, STA's step size of \trio{6}{8}{8} by definition requires tile size 384 for both Q and KV, which
        is not implementable with the current Blackwell FMHA, and therefore our implementation.
        We finally point out that with a speedup of approximately \Speedup{2.23} in the 91\% sparsity case,
        \textbf{GNA achieves the maximum speedup theoretically possible} for this exact level of sparsity
        (with respect to reduction in FLOPs).
    }
    \label{tab:vbench-summary}
\end{table*}

\begin{figure*}[ht]
    \centering
    \includegraphics[page=1, width=\textwidth]{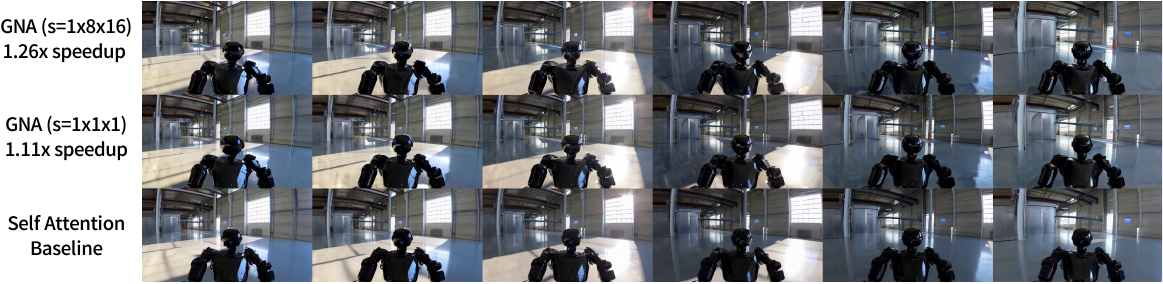}

    \includegraphics[page=2, width=\textwidth]{figures/video-samples/gna-samples.mini.pdf}

    \includegraphics[page=3, width=\textwidth]{figures/video-samples/gna-samples.mini.pdf}

    \includegraphics[page=4, width=\textwidth]{figures/video-samples/gna-samples.mini.pdf}
    \caption{
        Sample frames from videos generated by Cosmos-7B, with GNA introduced into the last 23 of the 35 diffusion steps.
        Window size is \trio{16}{32}{48} ($\approx$ 56\% sparsity).
        Speedup limit under this setting, with the same level of sparsity, is \Speedup{1.28}.
    }
    \label{fig:gna-cosmos7b}
\end{figure*}

We conduct experiments on the aforementioned video generation models: Cosmos-7B~\cite{agarwal2025cosmos} and HunyuanVideo~\cite{kong2024hunyuanvideo}.
Both use an isotropic Transformer architecture (DiT~\cite{peebles2023scalable} and MMDiT~\cite{esser2024scaling} respectively)
with full self attention.
In both cases, we generate 5 seconds of video at 720p.
In the case of Cosmos-7B, the token layout (DiT feature map shape) is \trio{16}{44}{80},
for which we use GNA with a window size of \trio{16}{32}{48}, which is approximately 56.4\% sparsity.
We found that going beyond this level of sparsity without further training/fine-tuning
we cannot maintain visual quality.
We tried both the best-performing GNA configuration (stride \trio{1}{8}{16}), and standard NA (stride \trio{1}{1}{1}).
Since the initial diffusion steps have a significant impact on the
overall structure of the generated video, we retain self attention for the first 12 diffusion steps and use GNA for the remaining
23 diffusion steps.
Because of this, GNA's share of the end-to-end workload decreases to approximately 39\%, which
in turn further limits achievable end-to-end speedup to approximately \Speedup{1.64}.
We present sample frames from some of the generated videos in
\cref{fig:gna-cosmos7b},
where we observe GNA can produce videos of comparable quality to that of the self attention baseline.

\begin{figure*}[ht]
    \centering
    \includegraphics[page=5, width=\textwidth]{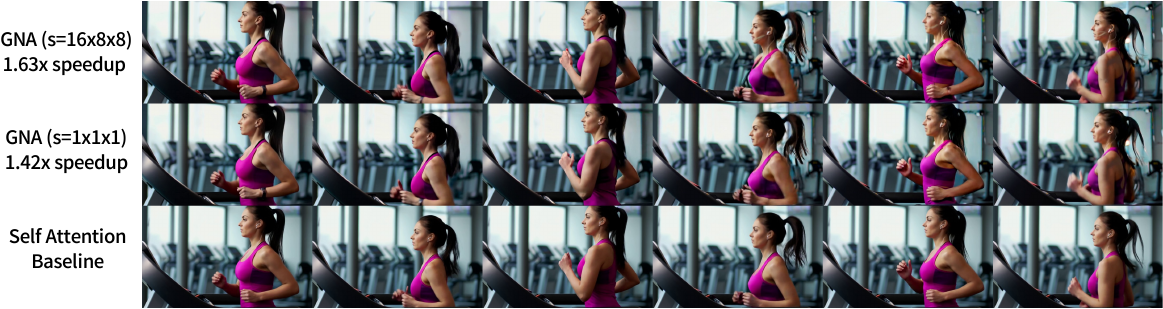}

    \includegraphics[page=6, width=\textwidth]{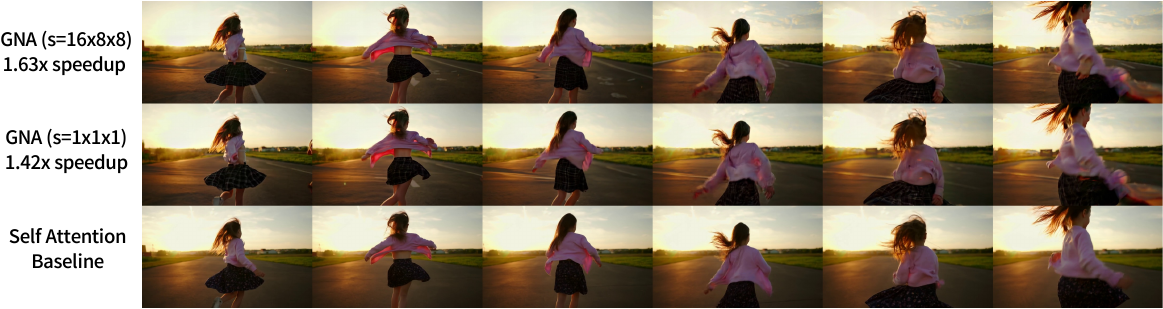}

    \includegraphics[page=7, width=\textwidth]{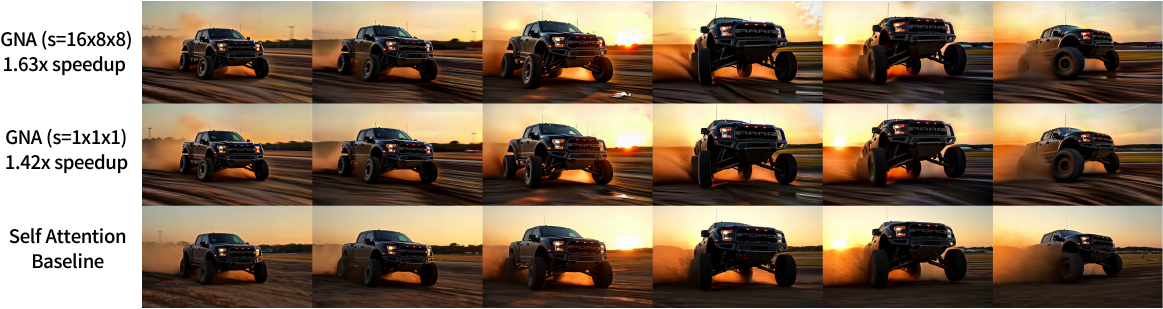}
    \caption{
        Sample frames from videos generated by HunyuanVideo, with GNA introduced into the last 35 of the 50 diffusion steps.
        Window size is \trio{18}{24}{24} ($\approx$ 91\% sparsity).
        Speedup limit under this setting, with the same level of sparsity, is \Speedup{1.63}.
    }
    \label{fig:gna-hunyuan}
\end{figure*}

In HunyuanVideo, the token layout (MMDiT feature map shape) is \trio{30}{48}{80}
for which we use GNA with a window size of \trio{18}{24}{24}, which is approximately 91\% sparsity,
and a stride of \trio{16}{8}{8}.
Similar to Cosmos, we retain self attention for some of the initial diffusion steps, which in this case we set to
15, and run GNA for the remaining 35 steps.
This decreases GNA's share of the end-to-end workload to approximately 42\%, which in turn
limits achievable end-to-end speedup to approximately \Speedup{1.72}.
We again try both the best-performing GNA configuration (stride \trio{16}{8}{8}), and standard NA (stride \trio{1}{1}{1}).
We present sample frames from output videos in
\cref{fig:gna-hunyuan}.

We additionally evaluate GNA with HunyuanVideo on VBench~\cite{huang2023vbench}.
In this case, we follow STA~\cite{zhang2025fast} by applying sparsity to all 50 steps.
However, we only report results without any additional training or fine tuning.
In \cref{tab:vbench-summary}, NA and GNA with 58.3\% sparsity achieve VBench scores of 83.24\% and 83.40\% respectively,
both of which are comparable to the self attention baseline at 83.08\%.
However, they can only achieve end-to-end speedups of \Speedup{1.11} and \Speedup{1.23} respectively.
When using higher sparsity of 91\%, the VBench score drops moderately to 82.36\% and 82.04\%, but with more significant end-to-end
speedups of \Speedup{1.73} and \Speedup{2.27} respectively.

\subsection{Image Generation}

\begin{table*}[ht!]
    \centering
    \resizebox{1.0\textwidth}{!}{
    \begin{tabular}{l||cc||ccccccc||c|rr|r}
        \toprule
        & \multicolumn{2}{c||}{\textbf{HPD} $\uparrow$} & \multicolumn{7}{c||}{\textbf{GenEval} $\uparrow$} & \textbf{Runtime} & \multicolumn{3}{c}{\textbf{E2E Speedup} $\uparrow$} \\
        
        \textbf{Configuration} & \textbf{MAN‑IQA} & \textbf{QualiCLIP} &
        \textbf{Overall} & \textbf{Single} & \textbf{Two}  & \textbf{Counting} & \textbf{Colors}& \textbf{Position} & \textbf{Color} & 
        \textbf{on B200}$\downarrow$ & \multicolumn{2}{c|}{\textbf{Analytical}} & \textbf{Actual} \\
        
        &&&&\textbf{Object}&\textbf{Objects} & & & & \textbf{Attr.} &
        \textbf{(seconds)} & \textbf{FLOP-wise} & \nattensimbold{} & \\ 
        
        \midrule
        \textbf{SA} (baseline) & \textbf{0.3718} & 0.4235 & \underline{0.5750} & 0.9594 & 0.6313 & \textbf{0.5500} & \underline{0.7793} & 0.1450 & 0.3850 &
         129 & &  &  \\
        \midrule
        \rowcolor[HTML]{EFEFEF} \multicolumn{14}{l}{\textbf{\textit{GNA with window size = \duo{80}{80} (90\% sparsity)}}} \\
        \textbf{s=\duo{1}{1}}    & \underline{0.3467} & \underline{0.4243} & 0.5743 & 0.9500 & \textbf{0.6742} & \underline{0.5031} & 0.7686 & \textbf{0.1500} & 0.4000 &
        94 & \Speedup{1.46} & \Speedup{1.42} & \Speedup{1.37} \\
        \textbf{s=\duo{16}{1}}   & \underline{0.3467} & \underline{0.4243}  & 0.5742 & \underline{0.9625} & 0.6607 & 0.4873 & 0.7739 & \underline{0.1480} & \underline{0.4125} &
        92 & \Speedup{1.46} & \Speedup{1.45} & \Speedup{1.40} \\
        \textbf{s=\duo{16}{16}}  & 0.3462 & \textbf{0.4247} & \textbf{0.5785} & \textbf{0.9656} & \underline{0.6717} & 0.4969 & \textbf{0.7819} & 0.1400 & \textbf{0.4150} &
        89 & \Speedup{1.46} & \Speedup{1.46} & \Speedup{1.45} \\
        \bottomrule
    \end{tabular}
    }
    \caption{
        4K image generation results from FLUX-1.dev~\cite{flux2024,yu2025ultra} on
        MAN-IQA~\cite{yang2022maniqa}, QualiCLIP~\cite{agnolucci2024quality}, and GenEval~\cite{ghosh2023geneval}.
        All GNA configurations retain self attention in the first 9 (out of 28) diffiusion steps, and
        use window size \duo{80}{80}, which is approximately 90\% sparsity.
        Similar to \cref{tab:vbench-summary}, we report generation runtime on the B200, and two measures for expected
        speedup: one based on reduction in FLOPs, and the other based on \nattensim{}.
        We also report actual speedup based on GNA runtimes, which use our Blackwell FNA kernel.
    }
    \label{tab:flux-results}
\end{table*}

\begin{figure*}[ht]
    \centering
    \includegraphics[page=1, width=\textwidth]{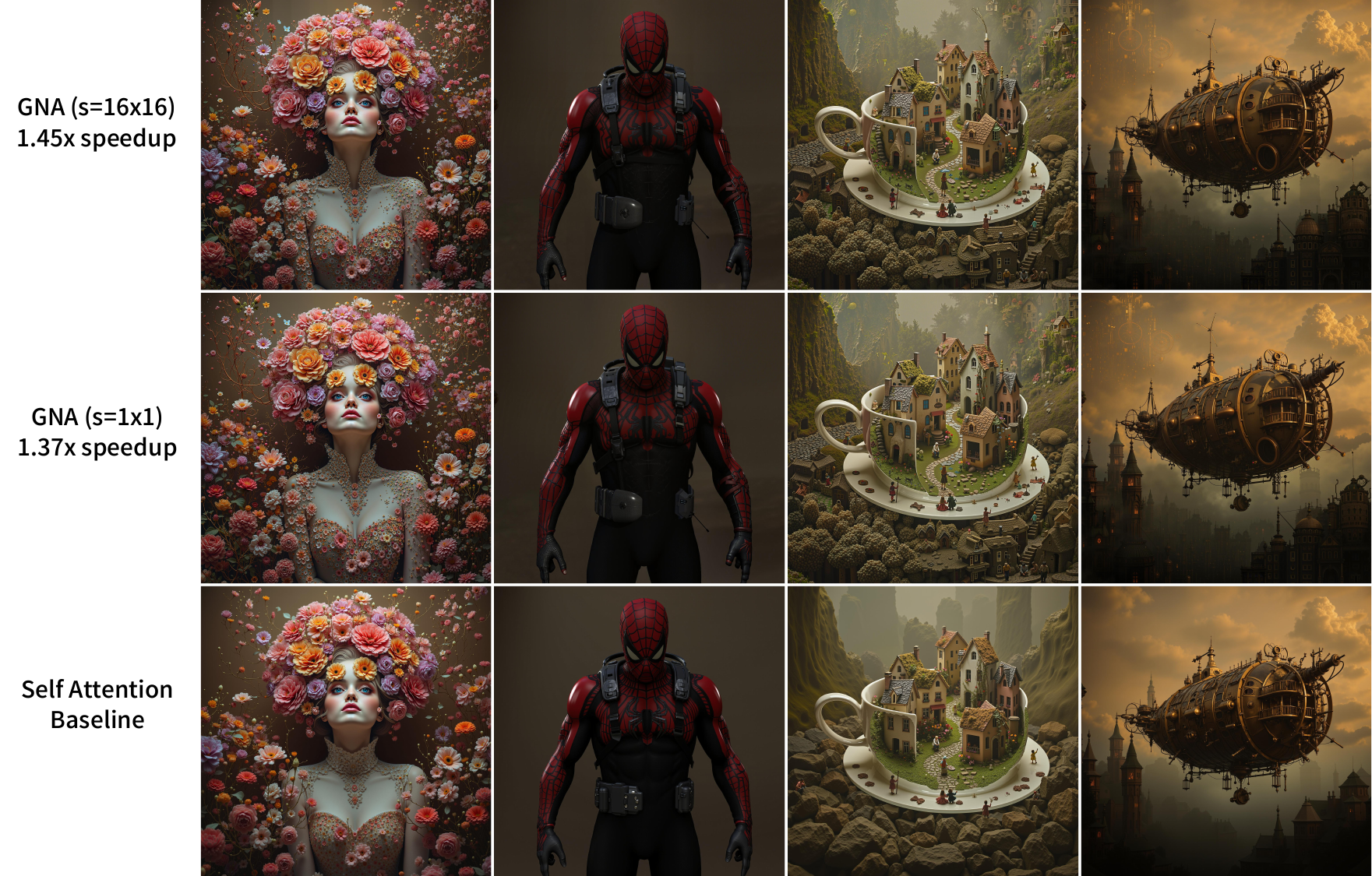}
    \caption{
        Images generated by ultra-resolution FLUX~\cite{flux2024,yu2025ultra} with GNA 
        introduced into the last 19 of the 28 diffusion steps.
        Window size is \duo{80}{80} ($\approx$ 90\% sparsity).
        Speedup limit under this setting, with the same level of sparsity, is \Speedup{1.46}.
    }
    \label{fig:flux-samples}
\end{figure*}

We conduct our experiments with image generation on FLUX-1.dev, which similar to the video models uses an
isotropic Transformer architecture (MMDiT~\cite{esser2024scaling}) with full self attention.
We only study generating 4K images, which result in \duo{256}{256} feature maps,
as smaller resolution workloads aren't as limited by the cost of self attention.
For generating 4K images, we use adapters from URAE~\cite{yu2025ultra}, as FLUX-1.dev does not natively
support 4K image generation.
We experiment with a window size of \duo{80}{80} (90.2\% sparsity), and strides \duo{1}{1}, \duo{16}{1}, \duo{16}{16},
which were the optimal choices from \nattensim{}.
Stride \duo{1}{1} is standard neighborhood attention, and stride \duo{16}{16} is perfectly block-sparse, which can
realize the full \Speedup{10.2} analytical op-level speedup.
For preserving quality and global structure, we follow our video generation experiments, in which we retain self attention
in the first few diffusion steps, which in this case we set to the first 9 out of the 28 diffusion steps.
This shrinks the share of GNA in the end-to-end workload to approximately 35\%,
and by extension the upper-bound end-to-end improvement to \Speedup{1.46}.
However, we successfully realize almost all of that with a speedup of \Speedup{1.45} over the self attention baseline.
In terms of quality, we observe very few notable differences when using GNA, even with a high stride of \duo{16}{16},
as illustrated in \cref{fig:flux-samples}.
We also present some quantitative metrics for these configurations.
Following URAE~\cite{yu2025ultra}, we evaluate them on MAN-IQA~\cite{yang2022maniqa}
and QualiCLIP~\cite{agnolucci2024quality} by generating images with prompts from the HPDv2~\cite{wu2023human} test dataset.
We additionally evaluate text-to-image alignment using GenEval~\cite{ghosh2023geneval}, which is
an object-focused evaluation benchmark. This benchmark is comprised of
various categories, each responsible for measuring a different compositional property, such as color, positioning,
attribute binding, object count, and the like.
We report the results of all three benchmarks in \cref{tab:flux-results}.
In summary, we observe FLUX-1.dev with GNA to be on par with the self attention baseline in terms of quality, while offering
up to 45\% end-to-end speedup on the B200.

\section{Conclusion}
Sliding window and blocked attention are two extremes of locality-focused sparse attention methods, with the former
enjoying inductive biases such as translational
equivariance~\cite{parmar2018image,ramachandran2019stand,vaswani2021scaling,hassani2023neighborhood} and arguably potential
for higher quality and expressiveness, and the latter potentially more efficient and trivial to implement.

In this paper, we extend the neighborhood attention family of patterns, which were already flexible in terms of
choice for window size, dilation, and causal masking, and add a new ``stride'' parameter. Just as in convolution, this parameter
adds a delay step to the sliding window effect, allowing it to implement many existing sparse attention methods.
Those include, but are not limited to, blocked local attention~\cite{vaswani2021scaling}, blocked / window self attention~\cite{liu2021swin},
and sliding tile attention~\cite{zhang2025fast}, in addition to the existing sliding window~\cite{parmar2018image,ramachandran2019stand} and
neighborhood attention~\cite{hassani2023neighborhood} methods.
While the concept of a delay step in sliding window attention is by no means new~\cite{vaswani2021scaling}, we revisit it
for a different reason compared to the original work.
While Vaswani et al.~\cite{vaswani2021scaling} were concerned with the cost of explicit memory operations, the focus of this
work is maximizing speedup from these methods to the point of being proportional to the level of sparsity (i.e. \Speedup{10} speedup
from 90\% sparsity).

We created an analytical model for GNA, called \nattensim{}, which can simulate tiling behavior under various design choices,
and compute the number of KV tiles visited per query tile, through which we compute more reliable upper-bound speedups for different
configurations under GNA.
These measures help quantitatively compare the upper-bound performance of different approaches without being biased by differences
in implementation.
In addition, we find that many GNA configurations exist that are perfectly block-sparse, under which speedup proportional to FLOPs
is possible, and they do not necessarily fit the definition of either blocked attention or sliding tile attention.

We further implement GNA on top of CUTLASS FMHA for the Blackwell architecture, which can achieve an effective 1.2 petaFLOPs/s
with FP16 and 1.7 petaFLOPs/s with FP8, and show that specifically in the case of perfectly block-sparse configurations it can
fully realize the analytical speedup computed by \nattensim{}.
We also highlight 3 potential applications for GNA, all of which we confirm spend the majority of their workload on self attention,
and report end-to-end speedups close to or matching the expected FLOP-wise speedup.
On Cosmos-7B~\cite{agarwal2025cosmos} and with 56\% sparsity introduced into 23 of 35 diffusion steps, we achieve
a speedup of \Speedup{1.26}, with FLOP-wise speedup being \Speedup{1.28}.
On HunyuanVideo~\cite{kong2024hunyuanvideo}, and with 91\% sparsity introduced into 35 of 50 diffusion steps,
we fully realize the FLOP-wise speedup of \Speedup{1.63}.
On FLUX-1.dev~\cite{flux2024}, and with 90\% sparsity introduced into 19 of 28 diffusion steps, we achieve
a speedup of \Speedup{1.45}, with FLOP-wise speedup being \Speedup{1.46}.
All three of the aforementioned configurations can still produce visually acceptable outputs, without any
further training or fine-tuning.

We hope that Generalized Neighborhood Attention can serve as a recipe for Speed-of-Light local attention beyond the use cases
and hardware studied in this paper.

\section{Future work}
Our new implementation can be further optimized by predicating fine-grained masking further in the event that
fully dense tiles exist in settings that are not perfectly block-sparse.
Since multi-dimensional tiling preserves spatial locality, if the intersection between the
neighborhoods of the first and last query in the Q tile spans entire KV tiles, fine-grained masking can be skipped.
Our current implementation does this for perfectly block-sparse configurations.
In addition to this, it is possible to performance-optimize the fine-grained mask logic itself.
These optimizations can further close the gap between the analytical speedup expected, and the actual speedup
achieved in cases that are only partially block-sparse, or not at all.

Token permutation and reverse permutation are also implemented naively with PyTorch, and barely utilize 1/8th of
the B200's peak memory bandwidth, and can be further optimized with specialized copy kernels and
activation memory management.
However, as noted earlier in the paper, the number of calls to these operations can be greatly reduced in certain
architectures, and with certain assumptions (i.e. isotropic Transformer architecture, and no dilation in GNA).

Other extensions can include transferring this design to other SOTA implementations for earlier architectures (i.e.
Flash Attention 3~\cite{shah2024flashattention} for Hopper and Flash Attention 2~\cite{dao2023flashattention}
for Ampere).

\section*{Acknowledgments}
We thank Alexander Kranias and Kai Wang for their feedback on this paper.

\noindent A.H. thanks Thomas M. Conte for inspiring some of the ideas in this project.

\clearpage

\clearpage

{\small
\bibliographystyle{ieee_fullname}
\bibliography{references}
}

\end{document}